\pdfoutput=1

\documentclass[11pt]{article}

\usepackage{authblk}

\usepackage[final]{acl}

\usepackage{times}
\usepackage{latexsym}

\usepackage{amsmath,amssymb,amsfonts}
\usepackage{graphicx}

\usepackage{booktabs}
\usepackage{textcomp}
\usepackage{xcolor}
\usepackage{tcolorbox}
\usepackage{colortbl}
\usepackage{bm}

\usepackage{url}            %
\usepackage{tabularx,booktabs}       %
\usepackage{nicefrac}       %
\usepackage{balance}
\usepackage{caption}
\usepackage{hyperref}

\usepackage{xspace}
\usepackage{color}
\usepackage{makecell}

\usepackage{algorithm}
\usepackage[noend]{algpseudocode}
\usepackage{algorithmicx}

\usepackage{lipsum}

\usepackage{xspace}
\usepackage{cleveref}
\usepackage{wrapfig}
\usepackage{enumitem}

\usepackage{framed}
\newenvironment{result}{\begin{framed}\centering\it}{\end{framed}}

\usepackage{multirow}
\def\BibTeX{{\rm B\kern-.05em{\sc i\kern-.025em b}\kern-.08em
    T\kern-.1667em\lower.7ex\hbox{E}\kern-.125emX}}

\usepackage{pifont}%

\let\emptyset\varnothing 

\algnewcommand{\LineComment}[1]{\State\(\triangleright\) #1}

\newcommand{\approach}{\textsc{KonTest}\xspace} %
\newcommand{\revise}[1]{\textcolor{black}{#1}}
\newcommand{\rrevise}[1]{\textcolor{black}{#1}}

\newcommand{\checknumber}[1]{\textcolor{black}{#1}}

\newcommand{\inspect}[1]{\textcolor{black}{#1}}
\newcommand{\aclRevise}[1]{\textcolor{black}{#1}}
\newcommand{\emnlpRevise}[1]{\textcolor{black}{#1}}

\newcommand{\gpt}{\texttt{GPT3.5}\xspace}
\newcommand{\lama}{\texttt{Llama2}\xspace}

\newcommand{\fal}{\texttt{Falcon}\xspace}
\newcommand{\gm}{\texttt{Gemini}\xspace}

\usepackage{bbding}    

\usepackage[T1]{fontenc}

\usepackage[utf8]{inputenc}

\usepackage{microtype}

\usepackage{inconsolata}

\title{Knowledge-based Consistency Testing of Large Language Models}

\author[1]{Sai Sathiesh Rajan\thanks{sai\_rajan@mymail.sutd.edu.sg}}
\author[2]{Ezekiel Soremekun\thanks{ezekiel.soremekun@rhul.ac.uk}}
\author[1]{Sudipta Chattopadhyay\thanks{sudipta\_chattopadhyay@sutd.edu.sg}}
\affil[1]{Singapore University of Technology and Design, Singapore}
\affil[2]{Royal Holloway, University of London, UK}

\begin{document}
\maketitle

\begin{abstract}
In this work, we systematically expose and measure the \emph{inconsistency} 
and \textit{knowledge gaps} of Large Language Models (LLMs).  
Specifically, we propose an automated testing framework (called \approach) which 
leverages a \textit{knowledge graph} to construct test cases. 
\approach probes and measures the inconsistencies in the LLM’s knowledge of the world
via a combination of 
semantically-equivalent queries  and test oracles (metamorphic or ontological oracle).  \aclRevise{
\approach further mitigates knowledge gaps 
via a weighted LLM model ensemble.}
Using four state-of-the-art LLMs (\fal, \gm,
\gpt, and \lama),  we show that 
 \approach generates \emnlpRevise{19.2\%} error inducing inputs 
\emnlpRevise{(1917 errors from 9979 test inputs)}.
It also reveals 
a 16.5\% 
knowledge gap 
across all tested LLMs.
\emnlpRevise{A mitigation method informed by \approach's test suite}
reduces LLM knowledge gap by 
32.48\%.
Our ablation study further 
shows that 
\texttt{GPT3.5}
is  
not suitable for knowledge-based consistency testing because it is 
only %
60\%-68\% effective in  
knowledge construction. 
\end{abstract}

\section{Introduction}
\label{sec:introduction}

Large language models (LLMs)
are being increasingly utilized in real-world 
applications.  
LLMs are powerful 
in solving many 
tasks,  but 
their \textit{reliability} remains 
a concern~\cite{qiu2023phenomenal}. 
This is alarming 
because 
inconsistent behaviors 
adversely affect critical 
downstream tasks and influence adoption. 

In this paper, we study the problem of 
assessing inconsistency 
in LLM behaviors. 
Previous works 
have demonstrated  the prevalence and severity of 
\textit{inconsistent} responses in LLMs~\cite{min2023beyond, berglund2023reversal,sallou2023breaking}. 
To address this challenge,  we conceptualize and design \approach\footnote{\revise{\approach means \textbf{\textsc{K}}nowledge-based 
		C\textbf{\textsc{on}}sistency \textbf{\textsc{Test}}ing}} -- a novel {\em test generation methodology to systematically 
discover consistency errors in LLMs and highlight their knowledge gaps}.

\begin{figure*}[tb!]
	\begin{center}
	\includegraphics[scale=0.48]{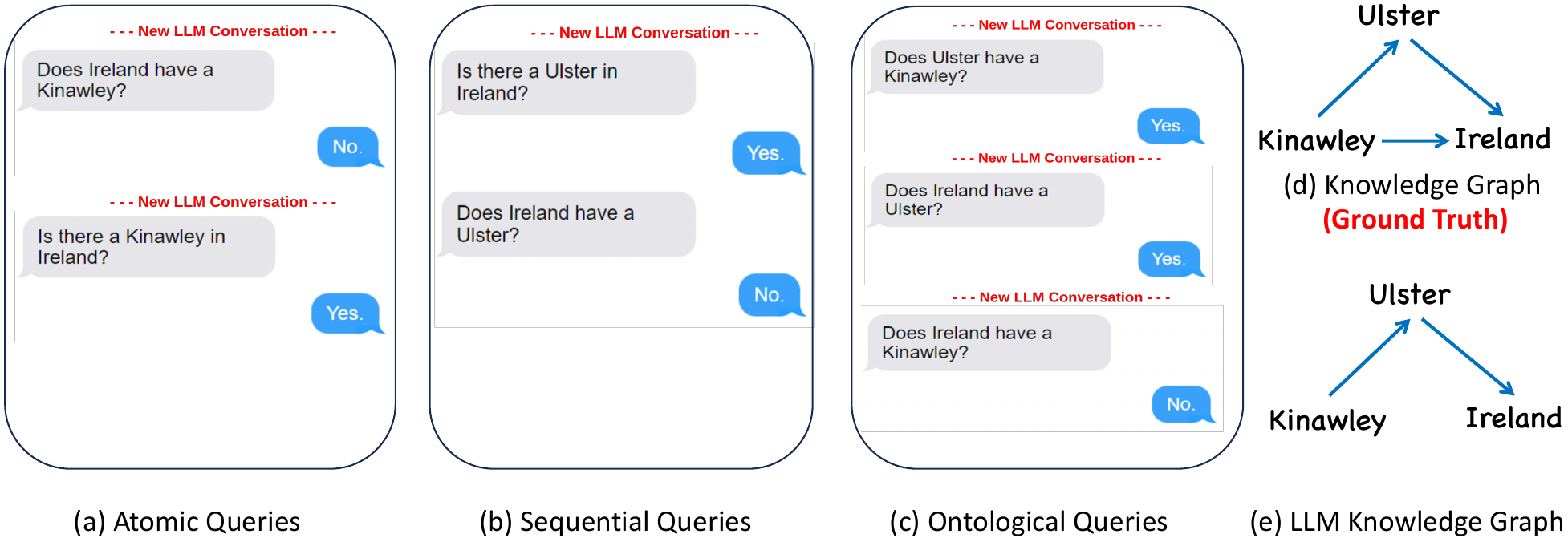}
	\end{center}
    \caption{Examples of queries generated and errors discovered by \approach in \gpt: Error reflected  (a) in two different conversations asking 
    	semantically equivalent questions 
    	({\bf Atomic} error in \autoref{tab:error-classes}) and   
    	(b) in conversation asking semantically equivalent questions %
    	in a sequence 
    	({\bf Sequential} error in \autoref{tab:error-classes}). (c) Example of  %
    	({\bf Ontological}) error in \autoref{tab:error-classes}, (d) the knowledge graph extracted for the considered 
    	entities i.e., Ulster, Kinawley and Ireland, (e) %
        SUT's knowledge showing that the LLM lacks the knowledge about {\em Kinawley is in Ireland}.}
    \label{fig:approach-example}
\end{figure*}

\autoref{fig:approach-example} shows examples of consistency errors and knowledge gap discovered by \approach in \gpt. 
These are errors in \gpt  
about the \texttt{place} (spatial) domain.
These errors 
may have adverse effects in critical areas (e.g., automobiles,  aeronautics) where spatial LLMs are 
deployed,  e.g.,  navigation systems -- MapBox's MapGPT~\cite{mapGPT}, 
LLM-Geo~\cite{li2023autonomous} and MapGPT~\cite{chen2024mapgpt}, L3MVN~\cite{Yu_2023}). 
\approach allows to automatically discover 
and expose such errors to end-users/developers.  
This is the \textit{first} step to 
enable their mitigation and repair for LLM improvement.

\approach leverages
a \textit{knowledge graph} for consistency testing (see \autoref{fig:approach-overview}).  
It first automatically extracts a set of entities and entity relationships  
from the knowledge graph.  
This is then used to 
systematically generate
(semantically-equivalent) \emnlpRevise{\textit{yes/no}} queries and create test cases for the \textit{subject LLM under test} (SUT) 
across various settings 
(e.g., 
atomic query vs. 
sequential 
query). 
\rrevise{The test case generation involves minimal, 
one-time effort for each type of entity relation considered (e.g., only two templates for 6730 test cases)
and such templates  can be reused for testing arbitrary LLMs.} 
\emnlpRevise{\approach's test suite can additionally be used to mitigate knowledge gaps}
by leveraging the 
likelihood of 
LLM inconsistencies 
to construct a weighted model ensemble.
To the best of our knowledge, {\em \approach is the first systematic approach 
for automatically generating  
consistency tests for assessing LLMs}.

Knowledge-based 
test generation provides a unique and systematic method for 
exploring the SUT's
knowledge and compute a \textit{test adequacy metric} for the SUT in terms of the covered entities and relations. 
More importantly, the responses from the SUT allow \approach to construct the {\em SUT's knowledge base} 
as a subset of the extracted knowledge graph 
and to concretely highlight the knowledge gap of the SUT.  This is valuable information for developers, it enables 
refining the model and improving its knowledge.

\approach differs from existing works both in its objective and in its unique approach for testing LLMs. 
Prior works have focused on other non-functional properties such as reasoning ability,  robustness and security~\cite{honarvar2023turbulence, wu2023reasoning,yang2023code}. 
Few works have demonstrated the importance of LLM consistency,  e.g.,  via 
self-consistency~\cite{min2023beyond} and reversal curse~\cite{berglund2023reversal}.
In contrast,
\approach employs \textit{automated test 
generation} to discover a larger scope of consistency errors.

This paper provides 
an overview of \approach (\autoref{sec:overview}),  and 
makes the following 
contributions:

\begin{enumerate}
\itemsep0em 	
\item  We present \approach, a novel approach to leverage \textit{knowledge graph} for checking the \textit{consistency} 
of LLM responses \revise{and measure their \textit{knowledge gaps}} (\autoref{sec:methodology}). 

\item We present the \textit{metamorphic} and \textit{ontological} oracles %
that allow to check 
consistency errors in LLM results  %
(\autoref{sec:methodology}). 

\item We implement \approach and evaluate it with %
\fal~\cite{falcon}, 
\gm~\cite{team2023gemini}, %
\gpt~\cite{gpt}, and \inspect{\lama~\cite{llama}.} 
\revise{
\approach revealed \emnlpRevise{19.2\%} erroneous inputs and exposed \inspect{an average knowledge gap of 16.5}\% %
(\autoref{sec:results}).}

\item
\emnlpRevise{We propose a technique that
mitigates knowledge gaps via a weighted model ensemble 
via the likelihood of LLM inconsistencies 
obtained from \approach's test suite.}
The mitigation technique 
reduces knowledge gaps by 32.48\% 
(\autoref{sec:results}).

\item We perform an ablation study by replacing each component 
of \approach with \gpt  using few-shot prompting. 
We found that 
\gpt constructs at most 68\% of the ground truth knowledge base.   
It exhibits up to \inspect{63.7}\% false positives in 
 error detection (\autoref{sec:results}). 

\end{enumerate}
After 
the related works (\autoref{sec:related-work}), we conclude 
in \autoref{sec:conclusion} and discuss limitations (\autoref{sec:threats}).

\section{Overview}
\label{sec:overview}

In this section, we outline the motivation behind our approach and present 
an illustrative example to demonstrate the overall process 
of our approach. 

\begin{figure}[h]
\centering
\includegraphics[scale=0.18]{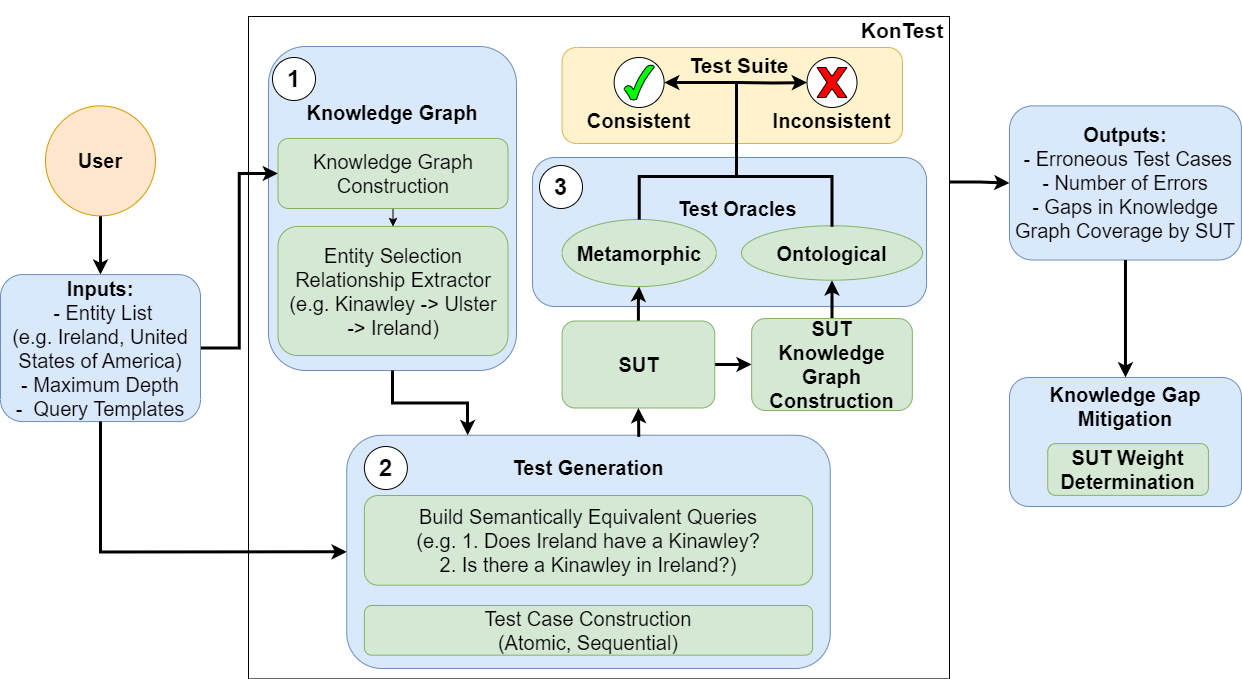}
\caption{Overall workflow of our approach (\approach)}
\label{fig:approach-overview}
\end{figure}

\begin{table*}[!bt]
	\centering
	\caption{Test cases generated and error types detected by \approach using metamorphic and ontological oracles.
	}
\label{tab:error-classes} 
{\small
\begin{tabular}{@{}ccc@{}}
	\toprule
	\multicolumn{1}{c}{\textbf{Generated Sentences}} & \textbf{Error Type} & \textbf{Oracle} \\ \midrule
	\multirow{6}{*}{\includegraphics[scale=0.1]{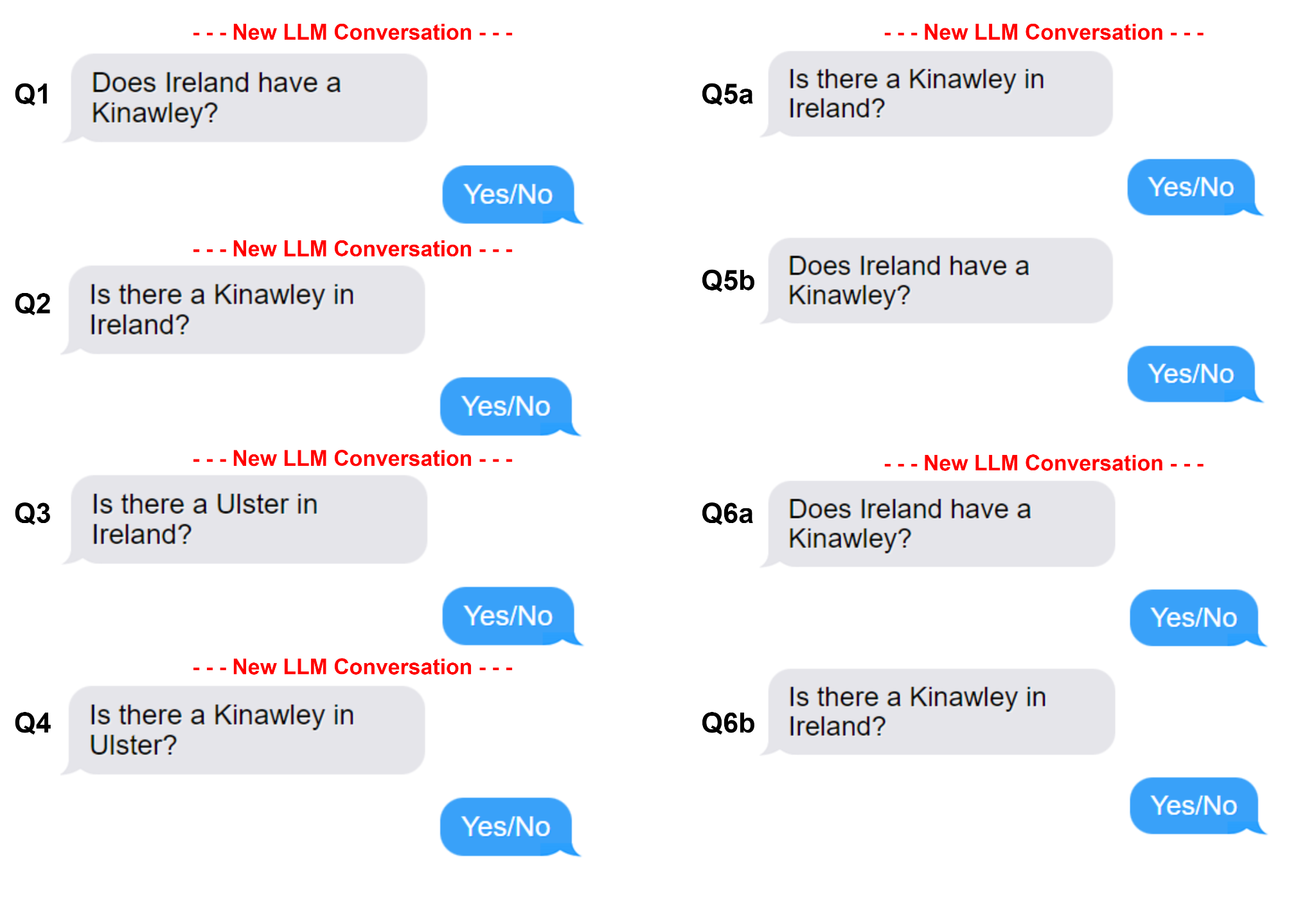}} & \begin{tabular}[c]{@{}c@{}} \\ {\tt atomic} \\ \end{tabular} & \begin{tabular}[c]{@{}c@{}} \\ LLM(Q1)=LLM(Q2) \\ \end{tabular} \\ \cmidrule(l){2-3} 
	& \begin{tabular}[c]{@{}c@{}} \\ {\tt sequential-intra} \\ \end{tabular} & \begin{tabular}[c]{@{}c@{}} \\ LLM(Q5a)=LLM(Q5b), \\ LLM(Q6a)=LLM(Q6b) \end{tabular} \\ \cmidrule(l){2-3} 
	& \begin{tabular}[c]{@{}c@{}} \\ {\tt sequential-inter} \\ \end{tabular} & \begin{tabular}[c]{@{}c@{}} \\ LLM(Q1)=LLM(Q5b), \\ LLM(Q2)=LLM(Q6b) \\ \end{tabular} \\ \cmidrule(l){2-3} 
	& \begin{tabular}[c]{@{}c@{}} \\ \\ {\tt Ontological} \\ \\ \end{tabular} & \begin{tabular}[c]{@{}c@{}} \\ (LLM(Q3)=LLM(Q4)=\textbf{Yes}) $\rightarrow$ \\ (LLM(Q2)=\textbf{Yes}) \\ \\ \end{tabular} \\
\end{tabular}}
\end{table*}

\smallskip\noindent
\textbf{Key Insight:} The key insight behind \approach is to leverage a knowledge graph for systematic
\revise{consistency testing} of LLMs. The 
knowledge graph serves multiple purposes in \approach: 
Firstly, 
entities and entity relations extracted from the knowledge graph 
allows \approach to systematically generate queries  
and 
construct 
test cases for validating the consistency of the subject LLM (SUT). 
The LLM's responses 
to these test cases 
allow \approach to build the knowledge (sub)-graph where the LLM behaves correctly. Secondly, knowledge-based test generation allows  \approach to report a 
\textit{test adequacy} metric for the 
SUT in terms of the entities and relations covered within the knowledge graph. Finally, 
this approach enables \approach to highlight sub-graph of the knowledge graph where the outputs from the SUT are logically {\em inconsistent}.  
This
enables practitioners to selectively focus on the knowledge gaps highlighted by \approach for improving 
the LLM (e.g., by fine-tuning).

\begin{figure}
	\includegraphics[scale=0.108]{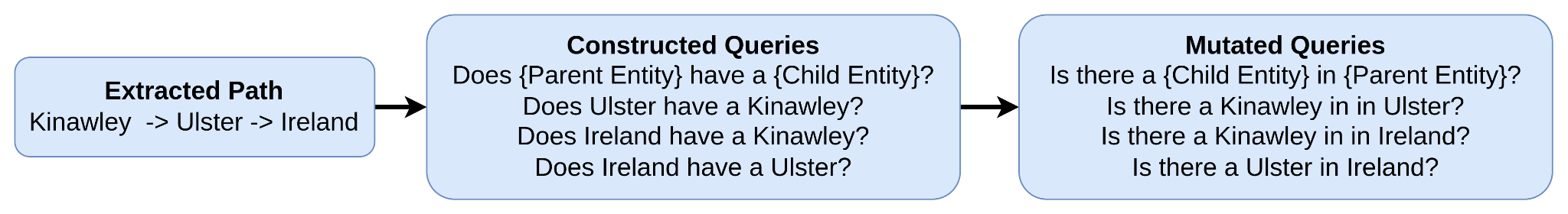}
	\caption{Example queries and \revise{templates} for the "Places" domain
in \approach
}
	\label{fig:Path-Query-Gen}
\end{figure}

\noindent
\textbf{Running Example:}
\autoref{fig:approach-overview} outlines our approach. \approach broadly comprises of three key components: 
\textcircled{1} knowledge graph construction (``Knowledge Graph'' in \autoref{fig:approach-overview}), 
\textcircled{2} test generation (``Test Generation'' in \autoref{fig:approach-overview}), and 
\textcircled{3} test oracles (``Test Oracles'' in \autoref{fig:approach-overview}). 
Given a set of entities (e.g., countries) and a maximum graph depth, 
\approach  locates the given entities in the knowledge graph 
(e.g., Wikidata knowledge graph~\cite{wikidata}), and constructs the associated knowledge graph before extracting the relevant entities and relationships. 
Given the relationships present in \autoref{fig:approach-example}(d), \approach constructs two semantically equivalent queries for each relation with the 
aid of a template (see \autoref{fig:Path-Query-Gen}) relating the two involved entities. The resultant queries are then fed to the SUT generating responses 
to both atomic and sequential queries. \autoref{fig:approach-example}(a) shows two such semantically equivalent queries, each of which was part of a new 
conversation with \gpt (aka ``\texttt{atomic}''), while  \autoref{fig:approach-example}(b) illustrates one such consistency error 
with inconsistent responses for semantically-equivalent queries that were within the same conversation (aka ``\texttt{sequential-inter}''). Finally,  
\autoref{fig:approach-example}(c) illustrates a series of queries, whose responses from \texttt{GPT3.5} collectively show inconsistent 
behavior (aka ``\texttt{ontological}'').   
Specifically, the positive responses to the first two queries imply that {\em Ireland does have a Kinawley}. However, 
due to the negative response to the third query, our ontological oracle reveals a consistency error. \aclRevise{We further note that such errors cannot be uncovered by counter-questioning LLMs.}
The relevant sub-graph of the knowledge graph 
for \gpt is shown in \autoref{fig:approach-example}(e). 
This clearly illustrates the knowledge gap 
(i.e., {\it Kinawley}$\nrightarrow${\it Ireland}) for \gpt. %

\section{Methodology}
\label{sec:methodology}

\subsection{Knowledge Graph based Test Generation}
\label{sec:test-gen}

\smallskip \noindent
\textbf{\textit{Knowledge Graph Construction}:} 
\approach allows the developer to specify the list of entities that are of interest. 
With these initial set of 
entities, \approach then queries an external source of knowledge to compute, up to a certain {\em depth}, all other 
related entities for the considered relation. Developers can also control this {\em depth}, allowing them to determine the 
degree to which the vicinity of the initial set of entities is explored. 
Once the initial knowledge base is constructed, %
\approach extracts the 
full path, $\mathbb{KG}$, associated with a leaf node of the knowledge base. %
This path is then used to guide the test generation process. \revise{For example, given the leaf entity {\em Kinawley}, as shown in 
	\autoref{fig:approach-example}, \approach extracts the $\mathbb{KG}$ for {\em Kinawley} as 
	$\mathit{Kinawley} \rightarrow \mathit{Ulster} \rightarrow \mathit{Ireland}$.}

\smallskip \noindent
\textbf{\textit{Test Generation}:} 
Given a knowledge path $\mathbb{KG}$, \approach leverages Algorithm~\ref{alg:query-gen} to exhaustively generate 
queries relating to all possible pairs of entities present in the path. 
To this end, \approach first finds the relation, $\mathbb{KG}_{rel}$, between a pair of nodes. 
\approach then generates a query, ${Query}$, using $\mathbb{KG}_{rel}$ with the aid of the template shown 
in \autoref{fig:Path-Query-Gen} (\cref{algl:query-line1} in  Algorithm~\ref{alg:query-gen}). 
\approach also generates the mutated query, ${Query_{M}}$, with another template 
(\cref{algl:query-line2} in  Algorithm~\ref{alg:query-gen}). These templates are %
dependent on the relation between the two nodes in question. 
For instance, \autoref{fig:Path-Query-Gen} 
shows the set of queries and mutated queries (i.e., $Query_{list}$ and $MutQuery_{list}$) generated from 
one particular path and along with the respective templates used to generate them. 

We note that developers can implement 
additional templates easily if they are interested in querying different relations. 
Moreover, creation 
of these templates is a one-time process and 
the cost incurred is minimal 
\emnlpRevise{(<10 LoC)}
since the number of different types of relationships 
is small in comparison to the number of entities. 
\emnlpRevise{We further note that our templates are directly applicable to other domains with similar relations. For instance, our templates for the "Places" domain can be reused for the "Food" domain since both domains have a "contains" type relation.}

\begin{algorithm}[t]
    \caption{KG-Based Query Generation}
    {\scriptsize
    \begin{algorithmic}[1]
        \Procedure{gen\_test}{$\mathbb{KG}$} 
        	\State $Query_{list}, MutQuery_{list} \gets \emptyset$
				\For {$\mathbb{KG}_{node1}$ in $\mathbb{KG}$}
        				\For {$\mathbb{KG}_{node2}$ in $\mathbb{KG}$}
        					\If{$\mathbb{KG}_{node1} \neq \mathbb{KG}_{node2}$}
        						\LineComment Finds relationship between $\mathbb{KG}_{node1}$ and $\mathbb{KG}_{node2}$
        						\State $\mathbb{KG}_{rel} \gets $ 
        						\textsf{Find\_Relation($\mathbb{KG}_{node1}, \mathbb{KG}_{node2}$)}
        						\LineComment Builds query using the relation 
        						($\mathbb{KG}_{node1}$, $\mathbb{KG}_{node2}$)
        						\State{$Query \gets $ \textsf{Gen\_Query($\mathbb{KG}_{node1}, \mathbb{KG}_{node2}, \mathbb{KG}_{rel}$)}  \label{algl:query-line1}}
        						\State{$Query_{list} \gets Query_{list} \cup \{{Query}\}$}
        						\LineComment Generates semantically equivalent query %
        						\State ${Query_{M}} \gets $ \textsf{Mut\_Gen(${Query}$)}  \label{algl:query-line2}
        						\State {$MutQuery_{list} \gets MutQuery_{list} \cup \{{Query_{M}}\}$}
        					\EndIf
        				\EndFor
					\EndFor
        	\State \Return $Query_{list}, MutQuery_{list}$
        \EndProcedure
    \end{algorithmic}
    }
    \label{alg:query-gen}
\end{algorithm}

\subsection{Test Oracles}
\label{sec:oracles}
\approach detects consistency errors in LLM %
via  %
a metamorphic oracle and an ontological oracle.

\smallskip \noindent
\textbf{\textit{Metamorphic Oracle}:} %
This oracle leverages 
both the original ($Query_{list}$) and mutated ($MutQuery_{list}$) query sets to create four unique conversations. 
These conversations are then fed into the target LLM (SUT). %
Concretely, \approach creates two conversations %
containing the answer to both (initial and mutated) \texttt{atomic} queries 
individually 
 (see the first column of ``Generated Sentences" 
in \autoref{tab:error-classes}) and two conversations where the queries are fed in a \texttt{sequential} manner (see the second 
column of ``Generated Sentences"  in \autoref{tab:error-classes}). Responses from these conversations are 
then evaluated using the consistency checker embodied in \approach %
to identify the erroneous test cases 
and the number of each error types (i.e., {\tt atomic}, {\tt sequential-intra} and {\tt sequential-inter}
in \autoref{tab:error-classes}).

\smallskip \noindent
\textbf{\textit{Ontological Oracle}:} 
Firstly, this oracle 
builds the SUT's 
knowledge graph. To this end, the generated queries from \approach are fed into the SUT and the target responses 
are recorded. 
Subsequently, given a list of such responses, 
\approach identifies the nodes 
involved in each response. These  
nodes are then added to the SUT's knowledge graph.
Next, if the query response is positive (i.e.,  correct), \approach introduces a directed edge 
between the nodes in the SUT's knowledge graph 
indicating that 
the respective relation exists between the two nodes being considered. It is important to note that the relations in question 
are not symmetric in nature, justifying the directed flavor of the edge. %
After constructing 
the SUT's knowledge graph, 
${Graph_{Gen}}$, \approach checks for ontological consistency errors with 
the aid of Algorithm~\ref{alg:graph-check}.

\begin{algorithm}[!bt]
	\caption{Graph Checker}
	{\scriptsize
		\begin{algorithmic}[1]
			\Procedure{Graph\_Checker}{${Graph_{Gen}}$}
			\State ${Err_{count}}, {Coverage_{count}} \gets \phi$
			\For {$Node_{1}$ in $Graph_{Gen}$}
			\For {$Node_{2}$ in $Graph_{Gen}$}
			\If{$Node_{1} \neq Node_{2}$}
			\LineComment Checks whether direct path exists between nodes.
			\State ${Path_{Dir}} \gets $ \textsf{Dir\_Path(${Node_{1}}, {Node_{2}}$)} \label{alg:graphCheck-line2}
			\If{$Path_{Dir} \neq TRUE$} \label{alg:grapCheck-line1}
			\LineComment Checks whether indirect path exists between nodes.
			\State ${Path_{InDir}} \gets $ \textsf{InDir\_Path(${Node_{1}}, {Node_{2}}$)}  \label{alg:graphCheck-line3}
			\If{$Path_{InDir} == TRUE$} \label{alg:graphCheck-line5}
			\State ${Err_{count}} \gets {Err_{count}} + 1$
			\EndIf
			\Else
			\State ${Coverage_{count}} \gets {Coverage_{count}} + 1$
			\EndIf
			\EndIf
			\EndFor
			\EndFor
			\State \Return $Err_{count}, {Coverage_{count}}$
			\EndProcedure
		\end{algorithmic}
	}
	\label{alg:graph-check}
\end{algorithm}

To check the consistency in SUT's knowledge 
graph,  Algorithm~\ref{alg:graph-check} first inspects whether a direct path exists between any pair 
of nodes ($Node_{1}, Node_{2}$) in ${Graph_{Gen}}$ (\cref{alg:graphCheck-line2} in 
Algorithm~\ref{alg:graph-check}). In the absence of such a direct path, when an indirect path exists 
between the same pair of nodes (\cref{alg:graphCheck-line5} in 
Algorithm~\ref{alg:graph-check}), \approach %
detects an ontological consistency error and updates the error count, ${Err_{count}}$. 
Intuitively, a direct path between two nodes indicates that the SUT responded positively 
to a question relating the two nodes. Similarly, the presence of an indirect path indicates that SUT knows %
that a positive relationship exists between the nodes albeit through more than one queries. 
As such, the SUT can be considered to be exhibiting an \textit{ontological inconsistency}.  
\revise{As an example, 
consider the ontological oracle illustrated in \autoref{tab:error-classes}. In this case, if an indirect path 
exists between nodes for {\em Kinawley} and {\em Ireland} via node {\em Ulster} (i.e., the responses to 
both {\bf Q3} and {\bf Q4} are positives from the SUT), then \approach concludes that a direct path 
{\em should also exist between Kinawley and Ireland}. Hence, the absence of a positive response for 
the respective query ({\bf Q2}) is indicated as an ontological error.}
As a byproduct of our ontological oracle, \approach computes the knowledge coverage ${Coverage_{count}}$  
for the SUT.

\subsection{Knowledge Gap Mitigation}
\label{sec:mitigation}
\emnlpRevise{We propose a weighted ensemble method that leverages the test suite generated by \approach to mitigate the discovered knowledge gaps.}
\emnlpRevise{The ensemble method}
uses the number of metamorphic errors found 
for each SUT on a per relation basis 
to inform its scoring system. A relation that 
induces $x$ errors is given a score of $5-x$ since each relation can maximally induce five errors. 
The cumulative SUT specific score, $\mathit{Score}_i$, is then found by summing the scores for each relation.
Next, 
\emnlpRevise{the relative weight, $\mathit{W}_i$, for each SUT in the set of SUTs ($\mathbb{SUT}$), is computed via the following equation:}
\begin{equation}
\label{eq:ensemble-weights}	
\mathit{W}_i  = \frac{\mathit{Score}_i}{\sum\mathit{Score}_i} \forall i \in \mathbb{SUT}
\end{equation}

\emnlpRevise{We then compute the final ensemble score} 
by first multiplying the relative weights with the results for each response and summing them. In particular, 
\emnlpRevise{we consider}
positive and negative responses to be one and zero respectively. 
If the ensemble score is above the midpoint (0.5),  the ensemble response for the relation is considered to be positive.  To determine the 
effectiveness 
of our weighting,  we compare it to \textit{simple majority voting}
--  an ensemble that does \textit{not} account
for the likelihood of inconsistencies for each SUT 
(\textit{see} RQ3).

\section{Evaluation Setup}
\label{sec:evaluation-setup}

To evaluate our approach (\approach), we pose the following \textit{research questions} (RQs):

\begin{itemize}[leftmargin=*]
\itemsep0em 
\item \textbf{RQ1 Effectiveness:} How \textit{effective} is \approach in
exposing consistency errors in LLMs?
\emnlpRevise{Are \approach results \textit{stable}?}

\item \textbf{RQ2 Knowledge Coverage:} What is the SUT's \textit{knowledge graph coverage}?
How much is the 
SUT \textit{knowledge gap} revealed by \approach?

\item \aclRevise{\textbf{RQ3 Knowledge Gap Mitigation:}
How \textit{effective} is 
\emnlpRevise{the proposed}
weighted ensemble technique in improving knowledge coverage? How does 
it compare 
to a simple majority voting? Does it generalize to a new template? 
}

\item \textbf{RQ4 Ablation Study with
\texttt{GPT3.5}:} %
Can the state-of-the-art LLM (\texttt{GPT3.5}) 
 perform the three main sub-tasks of \approach, i.e., knowledge 
construction, test generation and error detection?
\end{itemize}

\smallskip\noindent
\textbf{Knowledge Graph:} %
We rely on Wikidata’s knowledge graph~\cite{wikidata} and 
Wikidata's SPARQL query service to %
extract information pertinent to the two tested domains: {\em places} and {\em music}. 
\rrevise{We choose these domains as inconsistent knowledge about {\em places} may have adversarial 
consequences in navigation use cases of LLMs~\cite{mapGPT}. Concurrently,  
inconsistencies in the {\em music} domain may inaccurately capture the ownership of 
copyrighted materials (e.g., music albums).}
For the places domain, we take the list of countries ranked by (nominal) GDP per capita and select the top ten countries as our initial entity list~\cite{per-capita}. We then iteratively build our knowledge graph by finding ``{\em administrative divisions}" that are ``{\em located in the administrative territorial entity}" of the parent entity. For the music domain, we take the top 200 musical acts since 2000 as our initial entity list~\cite{music-acts}. We then find the set of ``{\em albums}" that have the parent entity as a ``{\em performer}". We also find the set of songs and singles that are ``{\em part of}" the album being considered. Once the graph is fully constructed, we randomly select 50 leaf nodes for each domain and extract the full path associated with the selected leaf nodes.

\smallskip\noindent
\textbf{Subject Programs and Query Construction:}
\autoref{tab:model-details} outlines features of the LLMs (SUTs) tested by \approach. We ensure that temperature for each query is set to zero where possible.
In addition, we provide the LLMs with an additional system command, ``{\em Be concise as possible. Answer with a yes or no response,}", 
before the start of any conversation. In the event that the LLM does not support a system command, we prepend the statement to the 
queries before feeding it to the LLM. For the queries themselves, we construct two sets of queries with the aid of a template relating the 
parent and child entities, as illustrated in \autoref{fig:Path-Query-Gen}.

\smallskip\noindent
\textbf{Test Generation and Evaluation:}
We provide each LLM with each of the queries present in the two sets (i.e., original and mutated) in an atomic manner. We then feed the queries in a sequential manner with queries from both sets being presented in turn. This is then repeated with the order reversed. To invoke the oracles, we only consider responses that begin with a {\em \bf yes} or {\em \bf no}. Other responses are classified as 
``{\em Invalid}". This allows us to 
determine 
the accuracy of the responses without penalizing LLMs that fail to respond appropriately.

\begin{table}[bt!]
	\caption{Model Details}
	\label{tab:model-details}
	\resizebox{\linewidth}{!}{
		\begin{tabular}{@{}ccccc@{}}
			\toprule
			\textbf{LLM} & \textbf{Lineage} & \textbf{Model} & \textbf{Size} & \textbf{Release Date} \\ \midrule
			GPT3.5 & GPT (OpenAI) & gpt-3.5-turbo & 175B & Mar 2023 \\
			Falcon & \begin{tabular}[c]{@{}c@{}}Falcon (TII) \\ Finetuned by Nomic\end{tabular} & gpt4all-falcon & 7B & Jun 2023 \\
			Llama2 & Llama (Meta) & Llama-2-7B & 7B & Jul 2023 \\
			Gemini & Gemini (Google) & gemini-pro & Unknown & Dec 2023 \\ \bottomrule
		\end{tabular}
	}
\end{table}

\smallskip\noindent \aclRevise{
\textbf{Knowledge Gap Mitigation:} 
We evaluate the effectiveness of 
\emnlpRevise{an ensemble method informed by \approach's test suite by determining the relative weights assigned to each SUT.}
We restrict the set of relations 
considered for this process 
such that a relation that yields an invalid answer in the query generation phase for any of the SUTs is not used in subsequent computations. We then perform five-fold cross validation on the remaining set of relations. Concretely, we hold out 20\% of the relations for evaluation and use the consistency errors associated with the rest of the relations in the computation of the weights. 
We introduce a third template to evaluate whether 
\emnlpRevise{our mitigation scheme}
generalizes to unseen templates.}

\smallskip\noindent
\textbf{Ablation Study:} 
We assess whether an LLM could conceivably be used to replicate each of the sub-tasks in our approach by testing its applicability on the places domain.
In particular, we chose to use \texttt{GPT3.5} for this due to its popularity and relatively high percentage (99.9\%) of 
non-exempted responses in our experiments. 
In each case described below, we provide 
\texttt{GPT3.5} with sample questions and answers (few shot prompts) to ensure that the outputs adhere to the format 
of \approach. %

To test the \textit{knowledge construction capability} of \texttt{GPT3.5}, 
we ask \texttt{GPT3.5} where each of the 50 leaf nodes (used to evaluate \approach) is located. %
We then check the accuracy of the response 
and we exempted 
responses that were not in the requested format.  %
To evaluate \texttt{GPT3.5}'s \textit{test generation capability}, we provide the relation we are concerned with and ask it 
to generate two semantically equivalent queries involving the two entities.
For ontological oracle, we provide \texttt{GPT3.5} with the full set of relations for each (knowledge) path and ask it to 
generate queries that expose inconsistencies in an LLM. 
For both oracles, %
if entity names and relations (as provided in the prompt) are not present in a generated query, then the test 
is considered invalid. 
For instance, ``{\em Is there a County of Capellen in County of Capellen?}" is  invalid since it 
does not correspond to the given knowledge relations.
For \textit{error detection},  we provide each set of queries along with their associated responses from each subject to \texttt{GPT3.5} and 
ask it to classify each set as being {\em Consistent} or {\em Inconsistent}. We also allow it to respond with {\em Invalid} when the 
SUT responses were exempted.
For the ontological oracle, we provide  \texttt{GPT3.5} with all the queries and responses relating to the entirety of a path and ask it to identify inconsistencies.
We note that \approach performs its check in Algorithm~\ref{alg:graph-check} using the same set of 
inputs provided to  \texttt{GPT3.5}.

\smallskip\noindent
\textbf{Implementation Details and Platforms:} 
\approach utilizes %
PyTorch 2.0,  CUDA 11.3 and the llm package. 
All experiments were conducted on a 
GCP 
VM with an N1 series machine, 
eight vCPUs, 30 GB of memory and 
one 
Nvidia T4 GPU.

\section{Evaluation Results}
\label{sec:results}

\begin{table*}[h]
\caption{Effectiveness of \approach using a \textit{knowledge graph}.}
\label{tab:meta-results}
\resizebox{\textwidth}{!}{
\begin{tabular}{@{}cccccccccccccc@{}}
\multicolumn{1}{l}{\multirow{3}{*}{\textbf{}}} & \multirow{3}{*}{\textbf{\begin{tabular}[c]{@{}c@{}}LLM \\(Subject)\end{tabular}}} & \multicolumn{6}{c}{\textbf{Valid Test Executions (\%)}} & \multicolumn{6}{c}{\textbf{Errors (\%)}} \\ \cmidrule(l){3-14} 
\multicolumn{1}{l}{} &  & \multicolumn{4}{c}{\textbf{Metamorphic}} & \multirow{2}{*}{\textbf{Ontological}} & \multirow{2}{*}{\textbf{All}} & \multicolumn{4}{c}{\textbf{Metamorphic}} & \multirow{2}{*}{\textbf{Ontological}} & \multirow{2}{*}{\textbf{All}} \\ \cmidrule(lr){3-6} \cmidrule(lr){9-12}
\multicolumn{1}{l}{} &  & \textbf{atomic} & \textbf{\begin{tabular}[c]{@{}c@{}}sequential \\-intra\end{tabular}} & \textbf{\begin{tabular}[c]{@{}c@{}}sequential \\-inter\end{tabular}} & \textbf{All Types} &  &  & \textbf{atomic} & \textbf{\begin{tabular}[c]{@{}c@{}}sequential \\-intra\end{tabular}} & \textbf{\begin{tabular}[c]{@{}c@{}}sequential \\-inter\end{tabular}} & \textbf{All Types} &  &  \\ \midrule 
\multirow{5}{*}{\rotatebox[origin=c]{90}{\textbf{Places}}} & \textbf{Falcon} & 254 (86.4) & 530 (90.1) & 538 (91.5) & 1322 (89.9) & 267 (91.4) & 1589 (90.2) & 50 (19.7) & 24 (4.5) & 152 (28.3) & 226 (17.1) & 25 (9.4) & 251 (15.8) \\
 & \textbf{Gemini} & 294 (100) & 588 (100) & 588 (100) & 1470 (100) & 292 (100) & 1762 (100) & 51 (17.3) & 6 (1.0) & 104 (17.7) & 161 (11.0) & 40 (13.7) & 201 (11.4) \\
 & \textbf{GPT3.5} & 293 (99.7) & 588 (100) & 587 (99.8) & 1468 (99.9) & 292 (100) & 1760 (99.9) & 61 (20.8) & 201 (34.2) & 138 (23.5) & 400 (27.2) & 17 (5.8) & 417 (23.7) \\
 & \textbf{Llama2} & 266 (90.5) & 543 (92.3) & 542 (92.2) & 1351 (91.9) & 268 (91.8) & 1619 (91.9) & 94 (35.3) & 52 (9.6) & 209 (38.6) & 355 (26.3) & 27 (10.1) & 382 (23.6) \\ \cmidrule(l){2-14} 
 & \textbf{Total} & 1107 (94.1) & 2249 (95.6) & 2255 (95.9) & 5611 (95.4) & 1119 (95.8) & 6730 (95.5) & 256 (23.1) & 283 (12.6) & 603 (26.7) & 1142 (20.4) & 109 (9.7) & 1251 (18.6) \\ \midrule
\multirow{5}{*}{\rotatebox[origin=c]{90}{\textbf{Music}}} & \textbf{Falcon} & 141 (96.6) & 287 (98.3) & 287 (98.3) & 715 (97.9) & 94 (97.9) & 809 (97.9) & 24 (17.0) & 38 (13.2) & 69 (24.0) & 131 (18.3) & 3 (3.2) & 134 (16.6) \\
 & \textbf{Gemini} & 146 (100) & 292 (100) & 292 (100) & 730 (100) & 96 (100) & 826 (100) & 39 (26.7) & 8 (2.7) & 77 (26.4) & 124 (17.0) & 7 (7.3) & 131 (15.9) \\
 & \textbf{GPT3.5} & 146 (100) & 292 (100) & 292 (100) & 730 (100) & 96 (100) & 826 (100) & 40 (27.4) & 85 (29.1) & 56 (19.2) & 181 (24.8) & 5 (5.2) & 186 (22.5) \\
 & \textbf{Llama2} & 137 (93.4) & 283 (96.9) & 276 (94.5) & 696 (95.3) & 92 (95.8) & 788 (95.4) & 46 (33.6) & 63 (22.3) & 97 (35.1) & 206 (29.6) & 9 (9.8) & 215 (27.3) \\ \cmidrule(l){2-14} 
 & \textbf{Total} & 570 (97.6) & 1154 (98.8) & 1147 (98.2) & 2871 (98.3) & 378 (98.4) & 3249 (98.3) & 149 (26.1) & 194 (16.8) & 299 (26.1) & 642 (22.4) & 24 (6.3) & 666 (20.5) \\ \midrule
\multicolumn{2}{c}{\textbf{Total}} & 1677 (95.3) & 3403 (96.7) & 3402 (96.6) & 8482 (96.4) & 1497 (96.5) & 9979 (96.4) & 405 (24.2) & 477 (14.0) & 902 (26.5) & 1784 (21.0) & 133 (8.9) & 1917 (19.2)
\end{tabular}
}
\end{table*}

\noindent
\textbf{RQ1 Effectiveness:} %
We found  that \textit{\approach} is effective in exposing consistency errors in LLMs: 
\emnlpRevise{19.2}\% of valid test executions result in consistency errors.
\autoref{tab:meta-results} also shows that metamorphic errors \emnlpRevise{(21.0\% = 1784/8482)} are more prevalent than ontological errors \emnlpRevise{(8.9\% = 133/1497).}  
We attribute \revise{the lower 
error rate of ontological errors/inputs to the high complexity of our ontological oracle 
(\textit{see} \autoref{tab:error-classes})}.  
\autoref{tab:meta-results} 
demonstrates that 
metamorphic errors are common across all LLMs. 
In particular, the tested 
LLMs are highly \textit{inconsistent} when asked the same question in a different manner (\textit{see} \autoref{tab:error-classes}).    
In addition, we observe that a large model size does not necessarily lead to a smaller error rate than a smaller model, e.g., 
\inspect{\gpt exhibits a metamorphic error rate of 27.2\% for the places entities as compared to a metamorphic error rate of 17.1\% for \texttt{Falcon}.}

\emnlpRevise{In addition, we validate the stability of our results by repeating our experiments four additional times. We find that the open-source models with frozen weights (\fal and \lama) yielded identical results when compared to the initial experiments. However, the closed-source models exhibited (up to 6\%) lower error rates on the subsequent iterations with \gm exhibiting a less than one percent change in error rate. We attribute this to changes in the underlying models~\cite{openai-deprecation} as the additional experiments were performed approximately eight (8) months after the initial experiments. Furthermore, we find that the error rates for all four subsequent iterations are fairly similar indicating that the results are consistent when repeated within a short duration of time (within a few days). In particular, we find that \gm results do not vary at all. Lastly, we also examine the stability of our results by executing 
both atomic and sequential queries separately.  For instance,
consider Q1 and Q6a (\autoref{tab:error-classes}) and the following additional oracle: LLM(Q1)=LLM(Q6a). We found that only GPT3.5 exhibited inconsistent results for this oracle and further noted that the absolute number of errors found was negligible (<1\%). These experiments demonstrate the stability of our results.}

\begin{result}
\approach effectively exposes consistency errors in LLMS: 
\emnlpRevise{19.2\%} of valid test executions %
exposed a metamorphic 
or ontological error in 
LLMs.  
\end{result}

\noindent
\textbf{RQ2 Knowledge Coverage:}
\autoref{tab:llm-cover} shows that 
\textit{the tested LLMs cover about 
four-fifth \inspect{(83.5\%)} 
of the tested knowledge graph across both templates.} 
We found that LLMs are particularly 
sensitive to the
query templates. For instance, \lama exhibits a 47.3\% gap in knowledge for the places entities for one template, but only exhibits a 12.9\% knowledge gap for the other. This suggests that 
LLMs
may respond differently to two different, but semantically equivalent,  input queries. 
We also observed that
the 
smallest model (\texttt{Falcon}) has the lowest knowledge gap for the places entities (3.1\%), 
while 
one of the biggest models (\gpt) has the highest knowledge gap (29.6\%). 
This implies that model size/complexity is not a good proxy for knowledge coverage/gap. We do, however, attribute the performance 
of \fal to its tendency to answer with a positive response regardless of the query. In addition, we also note that all queries posed to the 
subject LLMs are queries relating to an existing relation. 
These results show that \approach effectively reveals the knowledge gap in LLMs. 
We posit that knowledge coverage is a good 
criteria for estimating the underlying knowledge of an LLM.

\begin{result}
\approach 
effectively reveals knowledge gaps
in LLMs: It 
exposes an average knowledge gap of 16.5\% in the tested LLMs.
\end{result}

\begin{table}[!bt]
\caption{Knowledge Coverage and Gap in tested LLMs (Highest coverage or gap are marked in \textbf{bold text)}}
\label{tab:llm-cover}
\resizebox{\linewidth}{!}{
\begin{tabular}{@{}ccccccc@{}}
\multirow{2}{*}{} & \multirow{2}{*}{\textbf{\begin{tabular}[c]{@{}c@{}}LLM \\(Subject)\end{tabular}}} & \multicolumn{1}{l}{} & \multicolumn{3}{c}{\textbf{Knowledge Gap (\%)}} & \multirow{2}{*}{{\textbf{\begin{tabular}[c]{@{}c@{}}Overall \\Coverage (\%)\end{tabular}}}} \\
 &  & \textbf{Relations} & \textbf{Template 1} & \textbf{Template 2} & \textbf{Intersection} & \\ \midrule
\multirow{4}{*}{\rotatebox[origin=c]{90}{\textbf{Places}}} & \textbf{Falcon} & 294 & 12 (4.1) & 93 (31.6) & 9 (3.1) & \textbf{285 (96.9)} \\
 & \textbf{Gemini} & 294 & 60 (20.4) & 71 (24.1) & 40 (13.6) & 254 (86.4) \\
 & \textbf{GPT3.5} & 294 & \textbf{129 (43.9)} & 107 (36.4) & \textbf{87 (29.6)} & 207 (70.4) \\
 & \textbf{Llama2} & 294 & 38 (12.9) & \textbf{139 (47.3)} & 37 (12.6) & 257 (87.4) \\ \midrule
\multirow{4}{*}{\rotatebox[origin=c]{90}{\textbf{Music}}} & \textbf{Falcon} & 146 & 20 (13.7) & 15 (10.3) & 3 (2.1) & \textbf{143 (97.9)} \\
 & \textbf{Gemini} & 146 & 60 (41.1) & 45 (30.8) & 33 (22.6) & 113 (77.4) \\
 & \textbf{GPT3.5} & 146 & 63 (43.2) & 27 (18.5) & 25 (17.1) & 121 (82.9) \\
 & \textbf{Llama2} & 146 & \textbf{76 (52.1)} & \textbf{84 (57.5)} & \textbf{56 (38.4)} & 90 (61.6) \\ \midrule
\multicolumn{2}{c}{\textbf{Total}} & 1760 & 458 (26.0) & 581 (33.0) & 290 (16.5) & 1470 (83.5) \\
\end{tabular}
}
\end{table}

\begin{table*}[!tbh]
\caption{\aclRevise{
\centering Knowledge Gap Mitigation  results for \approach's weighted ensemble (``\approach''),  versus simple majority voting ensemble (``Majority'') and the initial \textit{average} knowledge gap found in the SUTs (``SUTs'').
Best reduction in knowledge gaps are in \textbf{bold text}. 
``\%Impr. '' means percentage improvement over the SUT. 
}}
\label{tab:ensemble-table}
\centering
\resizebox{\textwidth}{!}{
\begin{tabular}{@{}cccccccccccccc@{}}
\toprule
\multirow{3}{*}{\textbf{Domain}} & \multirow{3}{*}{\textbf{Relations}} & \multicolumn{12}{c}{\textbf{Knowledge Gap (\%)}} \\ \cmidrule(l){3-14} 
 &  & \multicolumn{3}{c}{\textbf{Template 1}} & \multicolumn{3}{c}{\textbf{Template 2}} & \multicolumn{3}{c}{\textbf{Template 3}} &  \multicolumn{3}{c}{\textbf{All (\%Impr.)}}\\ \cmidrule(l){3-14} 
 &  & \textbf{\approach} & \textbf{Majority} & \textbf{SUTs} & \textbf{\approach} & \textbf{Majority} & \textbf{SUTs} & \textbf{\approach} & \textbf{Majority} & \textbf{SUTs} & \textbf{\approach} & \textbf{Majority} & \textbf{SUTs} \\ \midrule
\textbf{Places} & 224 & \textbf{28 (12.5)} & 38 (17.0) & 38.25 (17.1) & \textbf{49 (21.9)} & 77 (34.4) & 65.75 (29.4) & \textbf{30 (13.4)} & 49 (21.9) & 41.5 (18.5) & \textbf{107 (26.46)} &  164 (-12.71) & 145.5 \\
\textbf{Music} & 132 & \textbf{39 (29.5) }& 60 (45.5) & 48.5 (36.7) & \textbf{15 (11.4)} & 44 (33.3) & 35.75 (27.1) & \textbf{24 (18.2)} & 55 (41.7) & 44.25 (33.5) & \textbf{78 (39.30)} & 159 (-23.74) & 128.5 \\ \midrule
\textbf{Total} & 356 & \textbf{67 (18.8)} & 98 (27.5) & 86.75 (24.4) & \textbf{64 (18.0)} & 121 (34.0) & 101.5 (28.5) & \textbf{54 (15.2)} & 104 (29.2) & 85.75 (24.1) & \textbf{185 (32.48)} & 323 (-17.88) & 274 \\ \bottomrule
\end{tabular}
}
\end{table*}

\begin{figure}[t]
	\centering
	\includegraphics[scale=0.75]{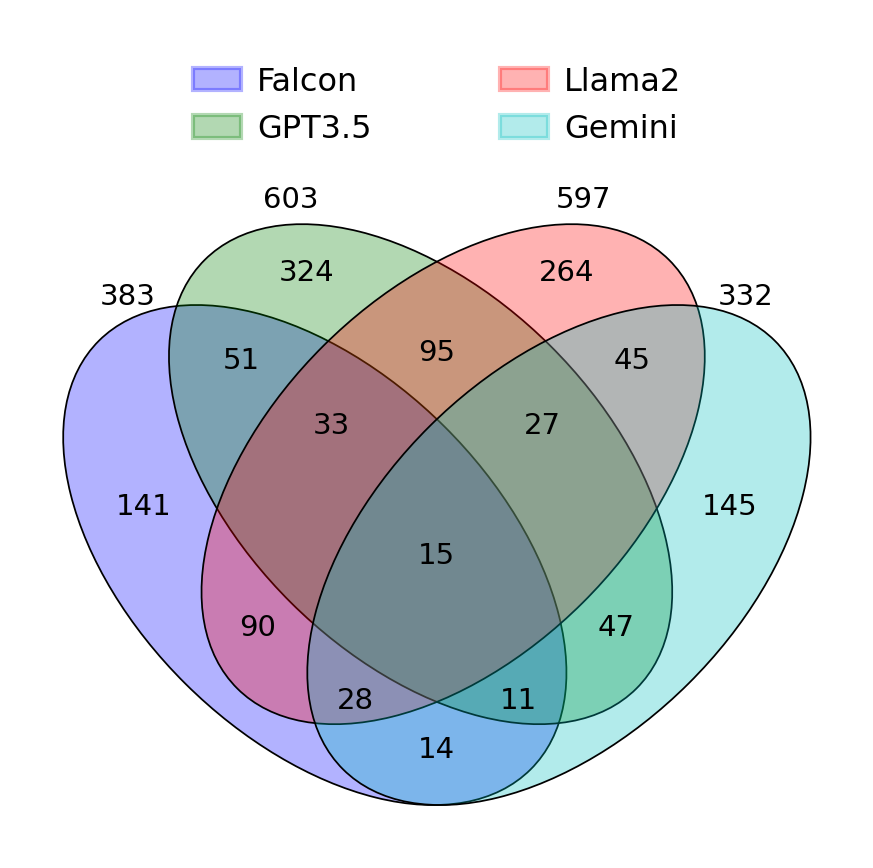}
	\caption{Venn Diagram of the consistency errors for each SUT.
	}
	\label{fig:venn-diagram}
\end{figure}

\noindent %
\textbf{RQ3 Knowledge Gap Mitigation:}
Results show that 
\emnlpRevise{our \textit{proposed mitigation technique reduces the SUT's knowledge gap by up to 39.30\%.}}
\autoref{tab:ensemble-table} shows that 
\emnlpRevise{our technique}
reduces the knowledge gap for all templates by 32.48\%.
We found 
that the simple majority voting ensemble 
worsens the SUTs' knowledge gap by up to 
23.74\%.  
We also observed that the mitigation performance of 
\emnlpRevise{our technique}
generalizes to an unseen template (template three (3)). While the 
\emnlpRevise{performance of our mitigation technique}
is slightly better on the new template than the original templates 
the simple majority performs worse on an unseen template.
\emnlpRevise{On the whole, we find that our weighted ensemble is 42.72\% ((323-185)/323) more effective than simple majority voting at reducing the knowledge gap.}
These results demonstrate the generalizability of 
\emnlpRevise{our proposed mitigation technique}
and the efficacy of our weighted ensemble.
We attribute the performance of 
\emnlpRevise{our technique to the}
likelihood and distribution of inconsistencies
\emnlpRevise{discovered by \approach (\autoref{fig:venn-diagram}).}
\emnlpRevise{\autoref{fig:venn-diagram} shows that}
only 1.1\% (15 out of 1330) of errors are found in all four SUTs while 65.7\% (874 out of 1330) of errors are unique to a single SUT.
\emnlpRevise{This suggests that an ensembling scheme, informed by the relative performance of the SUTs, reduces knowledge gap.}

\begin{result}
\aclRevise{
The 
\emnlpRevise{proposed}
ensemble method 
effectively reduces the SUT's knowledge gap by 
32.48\%.
}
\end{result}

\noindent
\textbf{RQ4 Ablation study with \gpt:}
We conduct an ablation study to investigate whether a state-of-the-art LLM (\gpt)
 is as
effective as 
\approach in performing 
its three main 
sub-tasks -- knowledge construction, test generation and 
error detection.

\begin{table}[!bt]
	\caption{\gpt efficacy in knowledge construction}
	\label{tab:prompt1}
	\resizebox{\linewidth}{!}{
		\begin{tabular}{@{}c|c|ccc|cc@{}}
			&  & \multicolumn{4}{c}{\textbf{Number of  LLM Responses (\%)}} \\
	& \textbf{Total} & \textbf{No Response} & \textbf{Exempted} & \textbf{Unexempted} & \textbf{Correct} & \textbf{Incorrect} \\ \midrule
	Nodes 
	& 50 & 0 (0)& 5 (10.0) & 45 (90.0) & 30 (60.0) & 15 (30.0) \\
	Edges 
	& 294 & 39 (13.3) & 30 (10.2) & 225 (76.5) & 200 (68.0) & 25 (8.5) \\
\end{tabular}
}
\end{table}

\noindent
\textbf{\textit{Knowledge Construction:}}
	\autoref{tab:prompt1} demonstrates that \gpt is not a reliable knowledge constructor. 
\gpt is only able to identify 68\% of the relations present in the original knowledge graph.
We believe this 
is because LLMs are generally intended to be a 
conversational engine,  rather than a knowledge database~\cite{pan2023unifying}.  
These results show the importance of knowledge graphs in \approach and suggest that LLMs are not a reliable replacement for knowledge graphs. 
\begin{result}
\gpt is an ineffective surrogate for a knowledge graph:
It
constructs only \inspect{68\%} of 
the ground-truth 
entity relations (edges). 
\end{result}

\begin{table}[]
\caption{Test generation effectiveness of \gpt and \approach on the places entities. ({\bf Meta.}: Metamorphic, {\bf Onto.}: Ontological)}
\label{tab:prompt-2-res}
\resizebox{\linewidth}{!}{
\begin{tabular}{@{}cccccccc@{}}
 & \multirow{2}{*}{\textbf{\begin{tabular}[c]{@{}c@{}}LLM \\(Subject)\end{tabular}}} & \multicolumn{3}{c}{\textbf{\begin{tabular}[c]{@{}c@{}}Valid Test \\Executions (\%)\end{tabular}}} & \multicolumn{3}{c}{\textbf{Errors (\%)}} \\
 & & \textbf{Meta.} & \textbf{Onto.} & \textbf{All} & \textbf{Meta.} & \textbf{Onto.} & \textbf{All} \\ \midrule
\multirow{4}{*}{\rotatebox[origin=c]{90}{GPT3.5}} & \textbf{Falcon} & 1322 (89.9) & 263 (93.9) & 1585 (90.6) & 232 (17.5) & 25 (9.5) & 257 (16.2) \\
 & \textbf{Gemini} & 1470 (100) & 280 (100) & 1750 (100) & 172 (11.7) & 41 (14.6) & 213 (12.2) \\
 & \textbf{Llama2} & 1351 (91.9)) & 264 (94.3) & 1615 (92.3) & 370 (27.4) & 27 (10.2) & 397 (24.6) \\ \cmidrule(l){2-8}
 & \textbf{Total} & 4143 (93.9) & 807 (96.1) & 4950 (94.3) & 774 (18.7) & 93 (11.5) & 867 (17.5) \\ \midrule
 & \textbf{\begin{tabular}[c]{@{}c@{}}\approach \\ w/o \gpt\end{tabular}} & 4143 (93.9) & 827 (94.4) & 4970 (94.0) & 742 (17.9) & 92 (11.1) & 834 (16.8)
\end{tabular}
}
\end{table}

\noindent
\textbf{\textit{Test Generation:}} %
Given a few (\checknumber{three}) query examples and the relations 
in a knowledge graph, 
\gpt is slightly more effective \inspect{(17.5\% vs 16.8\%)} than 
\approach
in test generation 
 (see \autoref{tab:prompt-2-res}).  
We also observe that this effectiveness persists for both the 
metamorphic \inspect{(18.7\% vs 17.9\%)} and the ontological oracle \inspect{(11.5\% vs 11.1\%).} 
We attribute the performance of \gpt to the effectiveness of few-shot prompting using 
the knowledge graph. This  guides \gpt to create tests similar to \approach. %
\begin{result}
		\gpt is slightly  (4.2\%) more effective than \approach in generating tests, when provided entity relations with few-shot prompting. 
\end{result}

\noindent
\textbf{\textit{Error Detection:}} 
\autoref{tab:oracle-task-breakdown} highlights 
that while \inspect{52.6\%} of the metamorphic errors detected by \gpt were detected correctly, approximately \inspect{40.1\%} of the errors were false positives~\footnote{Note that some errors detected by \gpt cannot be automatically validated by %
our oracles ({\bf UNK} in \autoref{tab:oracle-task-breakdown}), as it excludes responses that do not begin with {\bf yes/no}. }.
For ontological error detection, we provided \gpt all possible queries and corresponding responses for each knowledge path. \gpt was then asked to detect any inconsistency in these responses. Unlike \approach, \gpt is unable to detect multiple inconsistencies in a knowledge path. Hence, for a fair comparison,
we count ontological errors for \approach at the granularity of a knowledge path (i.e., at most one error per knowledge path). 
 	Since validating \gpt responses is not straightforward in this case, we manually validated the responses.
    We observed that the number of real ontological errors detected by \gpt is comparable to the 
    ontological errors detected by \approach, but it also had a high false positive rate with nearly \inspect{63.7\%} of errors being misclassified.
 
 \begin{result}
 	\revise{
 	 \gpt detects both metamorphic and ontological errors, but exhibits a false positive rate of up to 63.7\%.
}
 \end{result}

\section{Related Work}
\label{sec:related-work}

\begin{table}[]
\centering
\caption{Error Detection Effectiveness ({\bf TP/FP}: True/False positive, {\bf UNK}: Unknown)}
\label{tab:oracle-task-breakdown}
\resizebox{\columnwidth}{!}{
\begin{tabular}{@{}ccccccccc@{}}
\multirow{3}{*}{\textbf{Subject}} & \multicolumn{4}{c}{\textbf{Metamorphic Errors (\%)}} & \multicolumn{4}{c}{\textbf{Ontological Errors (\%)}} \\ \cmidrule(l){2-9} 
 & \multirow{2}{*}{\textbf{\begin{tabular}[c]{@{}c@{}}Kon- \\Test\end{tabular}}} & \multicolumn{3}{c}{\textbf{GPT3.5}} & \multicolumn{1}{l}{\textbf{}} & \multirow{2}{*}{\textbf{\begin{tabular}[c]{@{}c@{}}Kon- \\Test\end{tabular}}} & \multicolumn{2}{c}{\textbf{GPT3.5}} \\ \cmidrule(l){3-5} \cmidrule(l){8-9}
 & \multicolumn{1}{l}{} & \textbf{TP} & \textbf{FP} & \textbf{UNK} & \textbf{Paths} & \multicolumn{1}{l}{} & \textbf{TP} & \textbf{FP} \\ \midrule
\textbf{Falcon} & 226 & 208 (27.7) & 453 (60.4) & 89 (11.9) & 100 & 23 & 23 (33.3) & 46 (66.7) \\
\textbf{Gemini} & 161 & 161 (97.6) & 4 (2.4) & 0 (0) & 100 & 30 & 30 (54.5) & 25 (45.5) \\
\textbf{Llama2} & 355 & 352 (77.0) & 93 (20.4) & 12 (2.6) & 100 & 22 & 20 (26.0) & 57 (74.0) \\ \midrule
\textbf{Total} & 742 & 721 (52.6) & 550 (40.1) & 101 (7.4) & 300 & 75 & 73 (36.3) & 128 (63.7) \\
\end{tabular}
}
\end{table}

\noindent
\textbf{LLMs and Knowledge:} 
Pan et al.~\cite{pan2023unifying} presents %
techniques to combine LLMs and knowledge graphs to address 
their individual limitations. %
As an example,  Huang et al. ~\cite{huang2023adaptive} have demonstrated  
the feasibility of knowledge transfer to improve LLM's generalization ability in  software engineering tasks. %
Analogously, 
WEAVER~\cite{yang2023beyond} uses 
LLMs to generate knowledge bases, using which,  
requirements are extracted for testing models in real-world settings.  
GPT4GEO~\cite{gpt4geo} experimentally evaluates the capabilities and limitations  
of GPT-4 in geospatial domain (e.g., {\em places} entity), highlighting potential usage 
of GPT-4 in navigation. %
Unlike these proposed techniques,  
\approach %
employs knowledge graphs to 
expose inconsistencies and measure knowledge gaps in LLMs.

\noindent
\textbf{Testing and Analysis of LLMs:} 
Several researchers have surveyed the challenges and opportunities for testing and analysing LLMs~\cite{zhao2023survey, hou2023large, zheng2023towards, aleti2023software}. 
Researchers have also studied or proposed methods for testing and analyzing 
several quality properties of LLMs, including their 
reasoning ability~\cite{wu2023reasoning, qiu2023phenomenal}, 
 non-determinism~\cite{ouyang2023llm},   
 interpretability~\cite{palacio2023evaluating, rodriguez2023benchmarking},  
 robustness~\cite{zhu2023promptbench},  fairness~\cite{huang2023bias}, 
security concerns (e.g.,  privacy, memorization and backdoor attack)~\cite{yang2023code, staab2023beyond,huang2023composite}. 
In contrast 
to the aforementioned works,  
\approach studies the consistency and knowledge coverage of LLMs.

\section{Conclusion}
\label{sec:conclusion}

In this paper, we propose \approach where the key intuition is to distill (a subset of) facts from a knowledge graph, 
which are subsequently used to generate queries and formulate test cases for detecting a variety of consistency 
errors.  
Our evaluation reveals %
realistic consistency errors across state-of-the-art LLMs. Moreover, \approach opens opportunities to investigate 
LLMs through the lens of their knowledge gaps, which, in turn, is indicated as part of our framework. This  
helps designers and end users to understand and mitigate the effect of such knowledge gaps e.g., via prompt engineering 
or fine tuning. In future, we aim to investigate automated mitigation of consistency errors by leveraging the \approach framework. 
We provide our code and experimental data 
in the following: 
\begin{center}
	\url{https://github.com/sparkssss/KonTest}
\end{center}

\section{Limitations and Threats to Validity}
\label{sec:threats}

\noindent
\textbf{Construct Validity:} This relates to the metrics and measures employed in our experimental analysis.  To mitigate this threat, we have employed standard testing metrics such as the number/rate of generated inputs,  error-inducing inputs,  (knowledge) coverage and testing time. 
For automatic analysis of hundreds of responses,  
our analysis does not handle expressive, non-binary LLM responses. However, we mitigate this by employing system prompts to ensure the model provides binary responses and we discard non-binary responses (as invalid).%

\noindent
\textbf{Internal Validity:} This refers to the threat that our implementation of \approach performs its intended knowledge graph-based test generation.  We conduct several manual and automated tests, as well as inspection of sampled outcomes of \approach to ensure the correctness of our approach.  Our ablation study (\textbf{RQ4}) further allows us to probe the correctness of the sub-step of \approach versus using \texttt{GPT3.5}. \aclRevise{We experimentally verify that over 90\% of the entities we evaluate against existed in Wikidata prior to 2020. We also find that under 15\% of the errors found by \approach are linked to these entities. In addition, we also note that the SUT ought to answer the question in a consistent manner regardless of whether the entity existed prior to the corresponding knowledge cutoff date.}

\noindent
\textbf{External Validity:}
The main threat to external validity of this work is the generalizability of \approach and findings to LLMs,  
knowledge graphs, templates and entities beyond the ones used in this work. 
We mitigate the LLM generalizability threat by employing state-of-the-art off-the-shelf,  mature,  open 
model weights LLMs (\texttt{LLaMA2} and \texttt{FALCON}),  as well as commercial LLMs (such as \texttt{GPT3.5} and \texttt{Gemini}).  
\emnlpRevise{Similarly,}
our entity selection and template construction may be limited to our experimental setting. 
However, we \emnlpRevise{demonstrate the applicability of \approach by using two different domains with multiple relations and entity types.}
\emnlpRevise{However, we note that \approach might not be easily adapted to information that cannot be encapsulated within a knowledge graph.}
\emnlpRevise{In addition,  we employ few-shot prompting to conduct our ablation study, and our findings might not generalize to other prompting techniques.}
Finally,  \approach employs Wikidata, a well-maintained knowledge graph that is popularly used in both academia and industry~\cite{peters2019knowledge}.

\noindent
\textbf{LLM Stability and Correctness:}
Researchers have identified several stability concerns about LLMs~\cite{ouyang2023llm, fan2023large}, such as 
non-determinism,  randomness,  sensitivity to prompting methods,  and API/model updates. 
To mitigate these threats we performed several actions: First, 
we set the temperature of all models to zero (0), when possible~\cite{tempZero,ouyang2023llm}. 
Secondly, we reduce the risk of model updates by limiting our testing time 
(to about a day each per model) and checking for news of model updates 
before and after testing. 
Thirdly, we also use models with frozen weights (\fal and \lama) to reduce the non-determinism. 
To automatically validate model outcomes, %
we employ 
few-shot prompting, which has been shown to be effective for querying LLMs~\cite{deng2023large}.  
We examined whether LLMs are comparable to \approach (\textbf{RQ4})
using \texttt{GPT3.5}
since 
it produces the most valid responses (99.9\%) 
(\textit{see} \autoref{tab:meta-results}).

\noindent
\textbf{Knowledge Graph Completeness and Soundness:} 
We note that the knowledge graph is an incomplete snapshot of the real-world.  
In our evaluation, \approach only tests for facts (positive tests) derived 
from the knowledge graph. While it is also possible to use \approach to test for 
incorrect relation (negative test), the validation of such test results is challenging due 
to the incompleteness of the knowledge graph. Moreover, we only test 50 paths 
from this graph. However,  these concerns 
do not affect our findings, as we are certain about the errors found within the tested 
subset of the knowledge graph. Finally, knowledge of the world naturally evolves 
over time and the knowledge graphs do not evolve at the same pace~\cite{pan2023unifying}. 
To mitigate this, we use a widely used knowledge graph~\cite{wikidata}.

\section{Ethics Statement}
\label{sec:ethics}
We elucidate our ethics statement in this
section: (1) \textbf{Dataset}: We utilize data from Wikidata, a publicly available open knowledge base related to Wikipedia. Wikidata is licensed under the Creative Commons CC0 License.
(2) \textbf{Human Evaluations}: Our experiments do not involve human participants. 
(3) \textbf{Approach}: We test \approach with the aid of LLMs (both proprietary and free). We acknowledge that these models may give biased results due to their training data and methods. However, we restrict our queries to existing relations in the knowledge graph making it unlikely to raise ethical concerns. 
We limit ourselves to running inference on pre-trained models due to the numerous environmental concerns (energy and water expenditure) associated with training these LLMs.

\section{Acknowledgment}
\label{sec:funding}
This work was partially funded by grant number SMU-SUTD 2023\_02\_04 and the 
Singapore Ministry of Education (MOE) Present's Graduate Fellowship. Any 
opinions, findings, and conclusions or recommendations expressed in this material 
are those of the author(s) and do not reflect the view of the respective 
funding agencies.

\balance 

\bibliography{LLMorph}

\newpage

\end{document}